\documentclass[runningheads]{llncs}
\usepackage{algorithm,algorithmic,amsfonts,bm,booktabs,tikz,pgf,pgfplots,amsmath,soul,url,graphicx,amsmath,booktabs,booktabs,nicefrac,bm,colortbl,tikz,pgfplots,enumerate,xargs,optidef}
\usetikzlibrary{arrows,automata,circuits}
\usetikzlibrary{arrows,plotmarks,decorations.markings,trees,shapes}
\tikzstyle{nodo}=[ellipse,draw=black!100,fill=black!30,line width=.7pt,minimum width=1.2cm,minimum height=.7cm]
\tikzstyle{Qnodo}=[ellipse,draw=black!100,fill=black!10,line width=.7pt,minimum width=1.2cm,minimum height=.7cm]
\tikzstyle{arco}=[draw=black!80,line width=.7pt, postaction={decorate}, decoration={markings,mark=at position 1.0 with {\arrow[ draw=black!80,line width=.7pt]{>}}}]
\tikzstyle{decision} = [rectangle, draw, fill=black!100,text=white, text width=4.5em, text badly centered, node distance=3cm, minimum height=3em]
\tikzstyle{block} = [rectangle, draw, fill=blue!20, text width=5em, text centered, rounded corners, minimum height=3em]
\tikzstyle{line} = [draw, -latex']
\tikzstyle{cloud} = [draw, ellipse,fill=red!20, node distance=3cm, minimum height=2em]

\pgfplotsset{legend image with text/.style={
legend image code/.code={%
\node[anchor=center] at (0.3cm,0cm) {#1};}},}
\pgfplotsset{compat=1.13}
\begin{document}
\title{A New Score for Adaptive Tests\\ in Bayesian and Credal Networks}
\author{Alessandro Antonucci \and
Francesca Mangili \and \\
Claudio Bonesana \and
Giorgia Adorni}
\authorrunning{Antonucci et al.}
\institute{Istituto Dalle Molle di Studi sull'Intelligenza Artificiale, Lugano, Switzerland
\email{\{alessandro,francesca,claudio.bonesana,giorgia.adorni\}@idsia.ch}}
\maketitle
\begin{abstract}
A test is \emph{adaptive} when its sequence and number of questions is dynamically tuned on the basis of the estimated skills of the taker. Graphical models, such as Bayesian networks, are used for adaptive tests as they allow to model the uncertainty about the questions and the skills in an explainable fashion, especially when coping with multiple skills. A better elicitation of the uncertainty in the question/skills relations can be achieved by interval probabilities. This turns the model into a \emph{credal} network, thus making more challenging the inferential complexity of the queries required to select questions. This is especially the case for the information theoretic quantities used as \emph{scores} to drive the adaptive mechanism. We present an alternative family of scores, based on the mode of the posterior probabilities, and hence easier to explain. This makes considerably simpler the evaluation in the credal case, without significantly affecting the quality of the adaptive process. Numerical tests on synthetic and real-world data are used to support this claim.
\end{abstract}
\section{Introduction}\label{sec:intro}
A test or an exam can be naturally intended as a measurement process, with the questions acting as sensors measuring the skills of the test taker in a particular discipline. Such measurement is typically imperfect with the skills modelled as latent variables whose actual values cannot be revealed in a perfectly reliable way. The role of the questions, whose answers are regarded instead as manifest variables, is to reduce the uncertainty about the latent skills. Following this perspective, probabilistic models are an obvious framework to describe tests. Consider for instance the example in Figure \ref{fig:minicat}, where a Bayesian network evaluates the probability that the test taker knows how to multiply integers. In such framework making the test \emph{adaptive}, i.e., picking a next question on the basis of the current knowledge level of the test taker is also very natural. The information gain for the available questions might be used to select the question leading to the more informative results (e.g., according to Table \ref{tab:minicat}, $Q_1$ is more informative than $Q_2$ no matter what the answer is). This might also be done before the answer on the basis of expectations over the possible alternatives.

A critical point when coping with such approaches is to provide a realistic assessment for the probabilistic parameters associated with the modelling of the relations between the questions and the skills. Having to provide sharp numerical values for these probabilities might be difficult. As the skill is a latent quantity, complete data are not available for a statistical learning and a direct elicitation should be typically demanded to experts (e.g., a teacher). Yet, it might be not obvious to express such a domain knowledge by single numbers and a more robust elicitation, such as a probability interval (e.g., $P(Q_1=1|S_1=1)\in[0.85,0.95]$), might add realism and robustness to the modelling process \cite{hajek2012rationality}. With such generalized assessments of the parameters a Bayesian network simply becomes a \emph{credal} network \cite{antonucci2010d}. The counterpart of such increased realism is the higher computational complexity characterizing inference in credal networks \cite{maua2014jair}. This is an issue especially when coping with information theoretic measures such an information gain, whose computation in credal networks might lead to complex non-linear optimization tasks \cite{mangili2017b}. 

The goal of this paper is to investigate the potential of alternatives to the information-theoretic scores driving the question selection in adaptive tests based on directed graphical models, no matter whether these are Bayesian or credal networks. In particular, we consider a family of scores based on the (expected) mode of the posterior distributions over the skills. We show that, when coping with credal networks, the computation of these scores can be reduced to a simpler sequence of linear programming task. Moreover, we show that these scores benefit of better explainability properties, thus allowing for a more transparent process in the question selection.

\begin{figure}[htp!]
\centering
\begin{tikzpicture}[scale=1]
\node[] ()  at (5.5,0.5) {$P(Q_1=1|S=1)=0.9$};
\node[] ()  at (5.5,0.) {$P(Q_1=1|S=0)=0.3$};
\node[] ()  at (5.5,-0.5) {$P(Q_2=1|S=1)=0.6$};
\node[] ()  at (5.5,-1.) {$P(Q_2=1|S=0)=0.4$};
\node[nodo] (s)  at (0,0.5) {Knows multiplication ($S$)};
\node[Qnodo] (q1)  at (-2.,-1) {$10\times 5?$ ($Q_1$)};
\node[Qnodo] (q2)  at (+2,-1) {$13\times 14?$($Q_2$)};
\draw[arco] (s) -- (q1);
\draw[arco] (s) -- (q2);
\end{tikzpicture}
\caption{A Bayesian network over Boolean variables modelling a simple test to evaluate integer multiplication skill with two questions.}
\label{fig:minicat}
\end{figure}
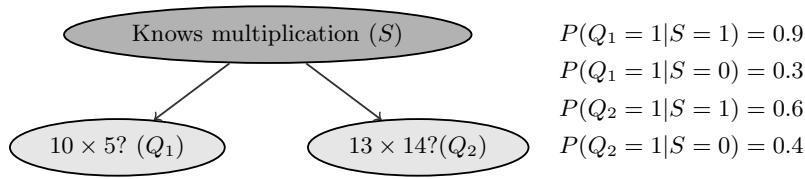
\begin{table}[htbp!]
\centering
\caption{Posterior probabilities of the skill after one or two questions in the test based on the Bayesian network in Figure \ref{fig:minicat}. A uniform prior over the skill is considered. Probabilities are regarded as grades and sorted from the lowest one. Bounds obtained with a perturbation $\epsilon=\pm 0.05$ of all the input parameters are also reported.}
\begin{tabular}{@{}ccrrr@{}}
\toprule
$Q_1$&$Q_2$&$P(S=1|q_1,q_2)$&
$\underline{P}(S=1|q_1,q_2)$&
$\overline{P}(S=1|q_1,q_2)$
\\
\midrule
$0$&$0$&$0.087$&$0.028$&$0.187$\\
$0$&$-$&$0.125$&$0.052$&$0.220$\\
$0$&$1$&$0.176$&$0.092$&$0.256$\\
$-$&$0$&$0.400$&$0.306$&$0.506$\\
$-$&$1$&$0.600$&$0.599$&$0.603$\\
$1$&$0$&$0.667$&$0.626$&$0.708$\\
$1$&$-$&$0.750$&$0.748$&$0.757$\\
$1$&$1$&$0.818$& $0.784$ &$0.852$\\
\bottomrule
\end{tabular}
\label{tab:minicat}
\end{table}
The paper is organized as follows. A critical discussion about the existing work in this area is in Section \ref{sec:work}. The necessary background material is reviewed in Section \ref{sec:background}. The adaptive testing concepts are introduced in Section \ref{sec:cat} and specialized to graphical models in \ref{sec:bncat}. The technical part of the paper is in Section \ref{sec:mode}, where the new scores are discussed and specialized to the credal case, while the experiments are in Section \ref{sec:exp}. Conclusions and outlooks are in Section \ref{sec:conc}.

\section{Related Work}\label{sec:work}
Modelling a test as a process relating latent and manifest variables since the classical \emph{item response theory} (IRT), that has been widely used even to implement adaptive sequences \cite{embretson2013item}. Despite its success related to the ease of implementation and inference, IRT might be inadequate when coping with multiple latent skills, especially when these are dependent. This moved researchers towards the area of probabilistic graphical models \cite{koller2009}, as practical tools to implement IRT in more complex setups \cite{mislevy1997graphical}. Eventually, Bayesian networks have been eventually identified as a suitable formalism to models tests, even behind the IRT framework \cite{vomlel2004bayesian}, this being especially the case for adaptive models \cite{vomlel2004building} and coached solving \cite{conati1997line}. In order to cope with latent skills, some authors successfully adopted EM approaches to these models \cite{plajner2020monotonicity}, this also involving the extreme situation of no ground truth information about the answers \cite{bachrach2012grade}.
As an alternative approach to the same issue, some authors considered relaxations of the Bayesian formalism, such as fuzzy models \cite{badaracco2013fuzzy} and imprecise probabilities \cite{mangili2017b}. The latter is the direction we consider here, but trying to overcome the computational limitations of that approach when coping with information-theoretic scores. This has some analogy with the approach in \cite{chen2015computer}, that is focused on the Bayesian case only, but whose score, based on the \emph{same-decision} problem, appears hard to be extended to the imprecise framework without affecting the computational complexity.

\section{Background on Bayesian and Credal Networks}\label{sec:background}
We denote variables by Latin uppercase letters, while using lowercase for their generic values, and calligraphic for the set of their possible values. Thus, $v \in \mathcal{V}$ is a possible value of $V$. Here we only consider discrete variables.\footnote{IRT uses instead with continuous skills. Yet, when coping probabilistic models, having discrete skill does no prevent evaluations to range over a continuous domain. E.g., see Table \ref{tab:minicat}, where the grade corresponds to a (continuous) probability.}

\subsection{Bayesian Networks}
A probability mass function (PMF) over $V$ is denoted as $P(V)$, while $P(v)$ is the probability assigned to state $v$. Given a function $f$ of $V$, its expectation with respect to $P(V)$ is $\mathbb{E}_P(f):=\sum_{v\in\mathcal{V}} P(v) f(v)$. The expectation of $-\log_b[P(V)]$ is called \emph{entropy} and denoted also as $H(X)$.\footnote{We set $0 \cdot \log_b 0 = 0$ to cope with zero probabilities.} We set $b:=|\mathcal{V}|$ to have the maximum of the entropy, achieved for uniform PMFs, equal to one.

Given joint PMF $P(U,V)$, the marginal PMF $P(V)$ is obtained by summing out the other variable, i.e., $P(v)=\sum_{u\in\mathcal{U}} P(u,v)$. Conditional PMFs such as $P(U|v)$ are similarly obtained by Bayes's rule, i.e., $P(u|v)=P(u,v)/P(v)$ provided that $P(v)>0$. Notation $P(U|V):=\{P(U|v)\}_{v\in\mathcal{V}}$ is used for such conditional probability table (CPT). The entropy of a conditional PMF is defined as in the unconditional case and denoted as $H(U|v)$. The conditional entropy is a weighted average of entropies of the conditional PMFs, i.e., $H(U|V):=\sum_{v\in\mathcal{V}} H(U|v) P(v)$. If $P(u,v)=P(u) P(v)$ for each $u\in\mathcal{U}$ and $v\in\mathcal{V}$, variables $U$ and $V$ are independent. Conditional formulations are also considered.

We assume the set of variables $\bm{V}:=(V_1,\ldots,V_r)$ to be in one-to-one correspondence with a directed acyclic graph $\mathcal{G}$. For each $V\in\bm{V}$, the parents of $V$, i.e., the predecessors of $V$ in  $\mathcal{G}$, are denoted as $\mathrm{Pa}_V$. Graph $\mathcal{G}$ together with the collection of CPTs $\{P(V|\mathrm{Pa}_V)\}_{V\in\bm{V}}$ provides a Bayesian network (BN) specification \cite{koller2009}. Under the Markov condition, i.e., every variable is conditionally independent of its non-descendants non-parents given its parents, a BN compactly defines a joint PMF $P(\bm{V})$ that factorizes as $P(\bm{v})=\prod_{V\in\bm{V}} P(v|\mathrm{pa}_V)$. Inference, intended as the computation of the posterior PMF of a single (queried) variables given some evidence about other variables, is in general NP-hard, but exact and approximate schemes are available (see \cite{koller2009} for details).

\subsection{Credal Sets and Credal Networks}
A set of PMFs over $V$ is denoted as $K(V)$ and called \emph{credal set} (CS). Expectations based on CSs are the bounds of the PMF expectations with respect to the CS. Thus $\underline{\mathbb{E}}[f]:=\inf_{P(V)\in K(V)} \mathbb{E}[f]$ and similarly for the supremum $\overline{\mathbb{E}}$. Expectations of events are in particular called lower and upper probabilities and denoted as $\underline{P}$ and $\overline{P}$. Notation $K(U|v)$ is used for a set of conditional CSs, while $K(U|V):=\{K(U|v)\}_{v\in\mathcal{V}}$ is a credal CPT (CCPT).

Analogously to a BN, a credal network (CN) is specified by graph $\mathcal{G}$ together with a family of CCPTs $\{K(V|\mathrm{Pa}_V)\}_{V\in\bm{V}}$ \cite{cozman2000}. A CN defines a joint CS $K(\bm{V})$ corresponding to all the joint PMFs induced by BNs whose CPTs are consistent with the CN CCPTs. For CNs, we intend inference as the computation of the lower and upper posterior probabilities. The task generalizes BN inference being therefore NP-hard, see  \cite{maua2014jair} for a deeper characterization. Yet, exact and approximate schemes are also available to practically compute inferences \cite{antonucci2013a}.

\section{Testing Algorithms}\label{sec:cat}
A typical test aims at evaluating the knowledge level of a test taker $\sigma$ on the basis of her answers to a number of questions. Let $\bm{Q}$ denote a repository of questions available to the instructor. The order and the number of questions picked from $\bm{Q}$ to be asked to $\sigma$ might not be defined in advance. We call \emph{testing algorithm} (TA) a procedure taking care of the selection of the sequence of questions asked to the test taker, and to decide when the test stops. Algorithm \ref{alg:ta} depicts a general TA scheme, with $\bm{e}$ denoting the array of the answers collected from test taker $\sigma$.

\begin{algorithm}[htp!]
\begin{algorithmic}[1]
\STATE $\bm{e}\gets\emptyset$
\WHILE{ {\bf not} $\tt{Stopping}(\bm{e})$}
\STATE $Q^* \gets \tt{Pick}(\bm{Q},\bm{e})$
\STATE $q^* \gets {\tt{Answer}}(Q^*,\sigma)$
\STATE $\bm{e} \gets \bm{e} \cup \{ Q^*=q^* \}$
\STATE $\bm{Q} \gets \bm{Q} \setminus \{ Q^*\}$
\ENDWHILE
\STATE {\bf return} $\tt{Evaluate}(\bm{e})$
\end{algorithmic}
\caption{General TA: given the profile $\sigma$ and repository $\bm{Q}$, an evaluation based on answers $\bm{e}$ is returned.\label{alg:ta}}
\end{algorithm}

Boolean function ${\tt Stopping}$ decides whether the test should end, this choice being possibly based on the previous answers in $\bm{e}$. Trivial stopping rules might be based on the number of questions asked to the test takes (${\tt Stopping}(\bm{e})=1$ if and only if $|\bm{e}|>n$) or on the number of correct answers provided that a maximum number of questions is not exceeded. Function ${\tt Pick}$ selects instead the question to be asked to the student from the repository $\bm{Q}$. A TA is called \emph{adaptive} when this function takes into account the previous answers $\bm{e}$. Trivial non-adaptive strategies might consist in randomly picking an element of $\bm{Q}$ or following a fixed order. Function ${\tt Answer}$ is simply collecting (or simulating) the answer of test taker $\sigma$ to a particular question $Q$. In our assumptions, this answer is not affected by the previous answers to other questions.\footnote{Generalized setups where the quality of the student answer is affected by the previous answers will be discussed at the end of the paper. This might include a \emph{fatigue} model negatively affecting the quality of the answers when many questions have been already answered as well as the presence of \emph{revealing} questions that might improve the quality of other answers \cite{laitusis2007examination}.} 

Finally, ${\tt Evaluate}$ is a function returning the overall judgement of the test (e.g., a numerical grade or a pass/fail Boolean) on the basis of all the answers collected after the test termination. Trivial examples of such functions are the percentage of correct answers or a Boolean that is true when a sufficient number of correct answers has been provided. Note also that in our assumptions the TA is \emph{exchangeable}, i.e., the stopping rule, the question finder and the evaluation function are invariant with respect to permutations in $\bm{e}$ \cite{sawatzky2016accuracy}. In other words, the same next question, the same evaluation and the same stopping decision is produced for any two students, who provided the same list of answers in two different orders.

A TA is supposed to achieve reliable evaluation of taker $\sigma$ from the answers $\bm{e}$. As each answer is individually assumed to improve such quality, asking all the questions, no matter the order because of the exchangeability assumption, is an obvious choice. Yet, this might be impractical (e.g., because of time limitations) or just provide an unnecessary burden to the test taker. The goal of a good TA is therefore to trade off the evaluation accuracy and the number of questions.\footnote{In some generalized setups, other elements such as a \emph{serendipity} in choice in order to avoid tedious sequences of questions might be also considered \cite{badran2019adaptive}.}

\section{Adaptive Testing in Bayesian and Credal Networks}\label{sec:bncat}
The general TA setup in Algorithm \ref{alg:ta} can be easily specialized to BNs as follows. First, we identify the profile $\sigma$ of the test taker with the actual states of a number of latent discrete variables, called \emph{skills}. Let $\bm{S}=\{S_i\}_{j=1}^n$ denote these skill variables, and $\bm{s}_\sigma$ the actual values of the skills for the taker. Skills are typically ordinal variables, whose states corresponds to increasing knowledge levels. Questions in $\bm{Q}$ are still described as manifest variables whose actual values are returned by the {\tt answer} function. This is achieved by a (possibly stochastic) function of the actual profile $\bm{s}_\sigma$. This reflects the taker perspective, while the teacher has clearly no access to $\bm{s}_\sigma$. As a remark, note that we might often coarsen the set of possible values $\mathcal{Q}$ for each $Q\in \bm{Q}$: for instance, a multiple choice question with three options might have a single right answer, the two other answers being indistinguishable from the evaluation point of view.\footnote{The case of \emph{abstention} to an answer and the consequent problem of modelling the incompleteness is a topic we do not consider here for the sake of conciseness. Yet, general approaches based on the ideas in \cite{marchetti2018a} could be easily adopted.} 
 
A joint PMF over the skills $\bm{S}$ and the questions $\bm{Q}$ is supposed to be available. In particular we assume this to correspond to a BN whose graph has the questions as leaf nodes. Thus, for each $Q\in\bm{Q}$, $Pa_Q \subseteq \bm{S}$ and we call $Pa_Q$ the \emph{scope} of question $Q$. Note that this assumption about the graph is simply reflecting a statement about the conditional independence between (the answer to) a question and all the other skills and questions given scope of the question. This basically means that the answers to other questions are not directly affecting the answer to a particular question, and this naturally follows from the exchangeability assumption.\footnote{Moving to other setups would not be really critical because of the separation properties of observed nodes in Bayesian and credal networks, see for instance \cite{antonucci2009,bolt}.}

As the available data are typically incomplete because of the latent nature of the skills, dedicated learning strategies, such as various form of constrained EM should be considered to train a BN from data. We refer the reader to the various contributions of Plajner and Vomlel in this field (e.g., \cite{plajner2020monotonicity}) for a complete discussion of that approach. Here we assume the BN quantification available.

In such a BN framework, ${\tt Stopping}(\bm{e})$ might be naturally based on an evaluation of the posterior PMF $P(\bm{S}|\bm{e})$, this being also the case for ${\tt Evaluate}$. Regarding the question selection, ${\tt Pick}$ might be similarly based on the (posterior) CPT $P(\bm{S}|Q,\bm{e})$, whose values for the different answers to $Q$ might be weighted by the marginal $P(Q|\bm{e})$. More specifically, entropies and conditional entropies are considered by Algorithm \ref{alg:bnta}, while the evaluation is based on a conditional expectation for a given utility function. 

\begin{algorithm}[htp!]
\begin{algorithmic}[1]
\STATE $\bm{e}=\emptyset$
\WHILE{$H(\bm{S}|\bm{e}) > H^*$}
\STATE $Q^* \gets \arg\max_{Q \in \bm{Q}} \left[ H(\bm{S}|\bm{e})-H(\bm{S}|Q,\bm{e}) \right]$
\STATE $q^* \gets {\tt{Answer}}(Q^*,\bm{s}_\sigma)$
\STATE $\bm{e} \gets \bm{e} \cup \{ Q^*=q^* \}$
\STATE $\bm{Q} \gets \bm{Q} \setminus \{ Q^*\}$
\ENDWHILE
\STATE {\bf return} $\mathbb{E}_{P(\bm{S}|\bm{e})}[f(\bm{S})]$
\end{algorithmic}
\caption{Information Theoretic TA in BN over the questions $\bm{Q}$ and the skills $\bm{S}$: given the student profile $\bm{s}_\sigma$, the algorithms returns an evaluation corresponding to the expectation of an evaluation function $f$ with respect to the posterior for the skills given the answers $\bm{e}$.\label{alg:bnta}}
\end{algorithm}

When no data are available for the BN training, elicitation techniques should be considered instead. As already discussed CNs might offer a better formalism to capture domain knowledge, especially by providing interval-valued probabilities instead of sharp values. If this is the case, a CN version of Algorithm \ref{alg:bnta} can be equivalently considered. Moving to CNs is almost the same, provided that bounds on the entropy are used instead for decisions. Yet, the price of such increased realism in the elicitation is the higher complexity characterizing inferences based on CNs. The work in \cite{mangili2017b} offers a critical discussion of those issues, that are only partially addressed by heuristic techniques used there to approximate such bounds. In the next section we consider an alternative approach to cope with CNs and adaptive TAs based on different scores used to select the questions.

\section{Coping with the Mode}\label{sec:mode}
Following \cite{wilcox1973indices}, we can regard the PMF entropy (and its conditional version) used by Algorithm \ref{alg:bnta} as an example of index of \emph{qualitative variation} (IQV). An IQV is just a normalized number that takes value zero for degenerate PMFs, one on uniform ones, being independent on the number of possible states (and samples for empirical models). The closer to uniform is the PMF, the higher is the index and vice versa.

In order to bypass the computational issues related to its application with CNs and the explainability limits with both BNs and CNs, we want to consider alternative IQVs to replace entropy in Algorithm \ref{alg:bnta}. Wilkox \emph{deviation from the mode} (DM) appears a sensible option. Given PMF $P(V)$, this corresponds to:
\begin{equation}
M(V):=1-\sum_{v\in\mathcal{V}} \frac{\max_{v'\in\mathcal{V}} P(v') - P(v)}{|\mathcal{V}|-1}\,.
\end{equation}
It is a trivial exercise to check that this is a proper IQV, with the same unimodal behaviour of the entropy. In terms of explainability, being a linear function of the modal probability, the numerical value of the DM offers a more transparent interpretation than the entropy. From a computational point of view, for both marginal and unconditional PMFs, both the entropy and the DM can be directly obtained from the probabilities of the singletons. 

The situation is different when computing the bounds of these quantities with respect to a CS. The bounds of $M(V)$ are obtained from the upper and lower probabilities of the singletons by simple algebra, i.e,
\begin{equation}
\overline{M}(V):=
\max_{P(V)\in K(V)} M(V):=\frac{|\mathcal{V}|-\max_{v'\in\mathcal{V}} \overline{P}(v')}{|\mathcal{V}|-1}\,,
\end{equation}
and analogously with the lower probabilities for $\underline{M}(V)$. Maximizing entropy requires instead a non-trivial, but convex, optimization. See for instance \cite{abellan2003maximum} for an iterative procedure to find such maximum when coping with CSs defined by probability intervals. The situation is even more critical for the minimization, that has been proved to be NP-hard in \cite{xiang2006estimating}.

The optimization becomes even more challenging for conditional entropies, there basically are mixtures of conditional entropies based on imprecise weights. Consequently, in \cite{mangili2017b}, only inner approximation for the upper bound have been derived. The situation is different for conditional DMs. The following result offers a feasible approach in a simplified setup, to be later extended to the general case.

\begin{theorem}\label{th:dm}
Under the setup of Section \ref{sec:bncat}, consider a CN with a single skill $S$ and a single question $Q$, that is a child of $S$. Let $K(S)$ and $K(Q|S)$ be the CCPTs of such CN. Let also $\mathcal{Q}=\{q^1,\ldots,q^n\}$ and $\mathcal{S}=\{s^1,\ldots,s^m\}$.
The upper conditional DM, i.e.,
\begin{equation}\label{eq:upperM}
\overline{M}(S|Q):=
|\mathcal{S}|-\max_{\substack{P(S)\in K(S)\\P(Q|S)\in K(Q|S)}} \sum_{i=1,\ldots,n} \left[ \max_{j=1,\ldots,m} P(s_j|q_i) \right] P(q_i)\,,
\end{equation}
whose normalizing denominator was omitted for the sake of brevity, is such that:
\begin{equation}\label{eq:upperM2}
\overline{M}(S|Q):=m-\max_{\substack{\hat{j}_i=1,\ldots,m\\i=1,\ldots,n}} \Omega(\hat{j}_1,\ldots,\hat{j}_n)\,,
\end{equation}
where $\Omega(\hat{j}_1,\ldots,\hat{j}_n)$ is the solution of the following linear programming task here below.
\begin{alignat}{3}
\max && \sum_j x_{ij}&&&\nonumber\\
\text{s.t.}\qquad&&\sum_{ij} x_{ij}&=1&& \label{c1}\\
&&x_{ij}&\geq 0&&\qquad \forall i,j \label{c2}\\
&&\sum_{i} x_{ij}&\geq \underline{P}(s_j)&&\qquad \forall j \label{c4}\\
&&\sum_{i} x_{ij}&\leq \underline{P}(s_j)&&\qquad \forall j \label{c5}\\
&&\underline{P}(q_i|s_j) \sum_{i} x_{ij}&\leq x_{ij}&&\qquad \forall i,j \label{c6}\\
&&\overline{P}(q_i|s_j) \sum_{i} x_{ij}&\geq x_{ij}&&\qquad \forall i,j
\label{c7}\\
&&x_{i\hat{j}_i}&\geq x_{ij}&&\qquad \forall i,j \label{c3}
\end{alignat}
Note that the bounds on the sums over the indexes and on the universal quantifiers
are also omitted for the sake of brevity. 
\begin{proof}
Equation \eqref{eq:upperM} rewrites as:
\begin{equation}\label{eq:firstopt}
\overline{M}(S|Q)
=m-\max_{\substack{P(S)\in K(S)\\P(Q|S)\in K(Q|S)}} \sum_{i=1}^n \left[ \max_{j=1,\ldots,m} P(s_j)P(q_i|s_j) \right]\,.
\end{equation}
Let us define the variables of such constrained optimization task as:
\begin{equation}
x_{ij}:=P(s_j) \cdot P(q_i|s_j)\,.
\end{equation}
for each $i=1,\ldots,n$ and $j=1,\ldots,m$. Let us show how the CCPT constraints can be easily reformulated with respect to such new variables by simply noticing that $x_{ij}=P(s_j,q_i)$, and hence $P(s_i)=\sum_i x_{ij}$ and $P(q_j|s_i)=x_{ij}/(\sum_{k} x_{kj})$. Consequently, the interval constraints on $P(S)$ corresponds to the linear constraints in Equations \eqref{c4} and \eqref{c5}. Similarly, for $P(Q|S)$, we obtain:
\begin{equation}\label{eq:fract}
\underline{P}(q_i|s_j) \leq \frac{x_{ij}}{\sum_k x_{kj}} \leq \overline{P}(q_i|s_j)\,,
\end{equation}
that easily gives the linear constraints in Equations \eqref{c6} and \eqref{c7}.
The non-negativity of the probabilities corresponds to Equation \eqref{c2}, while Equation \eqref{c1} gives the normalization of $P(S)$ and the normalization of $P(Q|S)$ is by construction. Equation \eqref{eq:firstopt} rewrites therefore as:
\begin{equation}\label{eq:lastopt}
\overline{M}(S|Q)
=m-\max_{ \{v_{ij}\}_{ij} \in \Gamma} \sum_{i} \max_{j} x_{ij} \,,
\end{equation}
where $\Gamma$ denotes the linear constraints in Equations \eqref{c1}-\eqref{c7}. If we set
\begin{equation}\label{eq:argmax}
\hat{j}_i:= \arg \max_j x_{ij}\,,
\end{equation}
Equation \eqref{eq:lastopt} rewrites as
\begin{equation}\label{eq:lastopt2}
\overline{M}(S|Q)
=\max_{ \{v_{ij}\}_{ij} \in \Gamma'} \sum_{i} x_{i\hat{j}_i} \,,
\end{equation}
where $\Gamma'$ are the constraints in $\Gamma$ with the additional (linear) constraints in Equation \eqref{c3}, that are implementing Equation \eqref{eq:argmax}.

The optimization on the right-hand side of Equation \eqref{eq:lastopt2} is not a linear programming task, as the values of the indexes $\hat{j}_i$ cannot be decided in advance being potentially different for different assignments of the optimization variables consistent with the constraints in $\Gamma$. Yet, we might address such optimization as a brute-force task with respect to all the possible assignation of the indexes $\hat{j}_i$. This is exactly what is done by Equation \eqref{eq:upperM2} where all the $m^n$ possible assignations are considered. This proves the thesis. \qed
\end{proof}
\end{theorem}

An analogous result with the linear programming tasks minimizing the same objective functions with exactly the same constraints allows to compute $\underline{M}(S|Q)$. 
The overall complexity is clearly $O(m^n)$ with $n:=|\mathcal{Q}|$. This means quadratic complexity for any test where only the difference between a wrong and a right answer is considered from an elicitation perspective, and tractable computations provided that the number of possible answers to the same question we distinguish is bounded by a small constant. Coping with multiple answers becomes trivial by means of the results in \cite{antonucci2009}, that allows to merge multiple observed children into a single one. Finally, the case of multiple skills might be similarly considered by using the marginal bounds of the single skills in Equations \eqref{c4} and \eqref{c5}. 

\section{Experiments}\label{sec:exp}
In this section we validate the ideas outlined in the previous section in order to check whether or not the DM can be used for TAs as a sensible alternative to information-theoretic scores such as the entropy. In the BN context, this is simply achieved by computing the necessary updated probabilities, while Theorem \ref{th:dm} is used instead for CNs.

\subsection{Single-Skill Experiments on Synthetic Data}
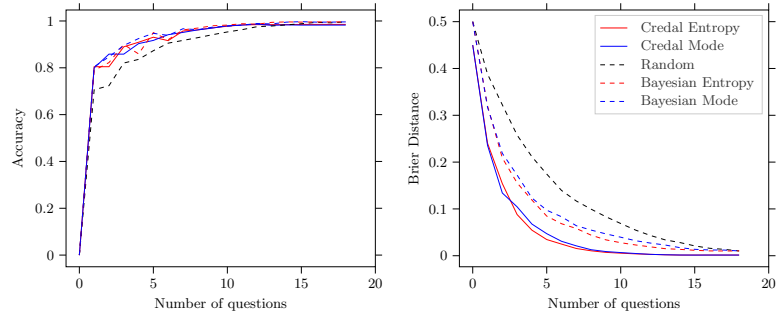
\begin{figure}[htp!]
\centering
\begin{tikzpicture}[scale=0.6]

\begin{axis}[
tick align=outside,
tick pos=both,
x grid style={white!69.0196078431373!black},
xlabel={Number of questions},
xmin=-0.9, xmax=20,
xtick style={color=black},
y grid style={white!69.0196078431373!black},
ylabel={Accuracy},
ymin=-0.049853515625, ymax=1.046923828125,
ytick style={color=black}
]
\addplot [semithick, red]
table {%
0 0
1 0.8046875
2 0.8046875
3 0.8896484375
4 0.9091796875
5 0.9306640625
6 0.916015625
7 0.95703125
8 0.9619140625
9 0.970703125
10 0.9775390625
11 0.9833984375
12 0.984375
13 0.984375
14 0.984375
15 0.9853515625
16 0.9853515625
17 0.9853515625
18 0.9853515625
};
\addplot [semithick, blue]
table {%
0 0
1 0.7998046875
2 0.8583984375
3 0.8583984375
4 0.9033203125
5 0.9169921875
6 0.94140625
7 0.951171875
8 0.9619140625
9 0.9697265625
10 0.9794921875
11 0.982421875
12 0.986328125
13 0.9814453125
14 0.9833984375
15 0.982421875
16 0.982421875
17 0.982421875
18 0.982421875
};
\addplot [semithick, black, dashed]
table {%
0 0
1 0.7060546875
2 0.7236328125
3 0.8203125
4 0.8359375
5 0.873046875
6 0.9052734375
7 0.9169921875
8 0.9287109375
9 0.94140625
10 0.953125
11 0.962890625
12 0.9755859375
13 0.978515625
14 0.9833984375
15 0.98828125
16 0.994140625
17 0.9921875
18 0.99609375
};
\addplot [semithick, red, dashed]
table {%
0 0
1 0.7890625
2 0.822265625
3 0.8994140625
4 0.859375
5 0.9501953125
6 0.919921875
7 0.9619140625
8 0.9697265625
9 0.9794921875
10 0.982421875
11 0.98828125
12 0.986328125
13 0.994140625
14 0.99609375
15 0.9951171875
16 0.99609375
17 0.99609375
18 0.99609375
};
\addplot [semithick, blue, dashed]
table {%
0 0
1 0.8046875
2 0.8466796875
3 0.8955078125
4 0.9248046875
5 0.947265625
6 0.9384765625
7 0.966796875
8 0.962890625
9 0.9736328125
10 0.9775390625
11 0.982421875
12 0.9892578125
13 0.9873046875
14 0.9951171875
15 0.9970703125
16 0.9951171875
17 0.9951171875
18 0.99609375
};
\end{axis}

\end{tikzpicture}
\begin{tikzpicture}[scale=0.6]
\begin{axis}[
legend cell align={left},
legend style={fill opacity=0.8, draw opacity=1, text opacity=1, draw=white!80!black},
tick align=outside,
tick pos=both,
x grid style={white!69.0196078431373!black},
xlabel={Number of questions},
xmin=-0.9, xmax=20,
xtick style={color=black},
y grid style={white!69.0196078431373!black},
ylabel={Brier Distance},
ymin=-0.0235694775390625, ymax=0.524931879882813,
ytick style={color=black}
]
\addplot [semithick, red]
table {%
0 0.449999999999992
1 0.240321093750003
2 0.154068750000002
3 0.0879601562499998
4 0.0545979492187504
5 0.0345669921874999
6 0.0248742187499998
7 0.0154591796875
8 0.0104958984375
9 0.00716503906250006
10 0.00544667968750006
11 0.00400634765625
12 0.00290869140625
13 0.00200585937499999
14 0.00137460937499999
15 0.00145478515624999
16 0.00144414062499999
17 0.00143984374999999
18 0.00143974609374999
};
\addlegendentry{Credal Entropy}
\addplot [semithick, blue]
table {%
0 0.449999999999992
1 0.235816308593751
2 0.133981542968751
3 0.1044369140625
4 0.0676275390624998
5 0.0471173828125
6 0.0308186523437499
7 0.0210914062499999
8 0.0128307617187499
9 0.00890488281249997
10 0.00674853515625
11 0.00476650390625
12 0.00295224609375002
13 0.00205888671875
14 0.00155673828125
15 0.00136240234375
16 0.00144453125
17 0.0014396484375
18 0.001439453125
};
\addlegendentry{Credal Mode}
\addplot [semithick, black, dashed]
table {%
0 0.5
1 0.38800205078125
2 0.32236396484375
3 0.25762578125
4 0.21144267578125
5 0.1745076171875
6 0.139484375
7 0.116849609375
8 0.09946396484375
9 0.0832636718750001
10 0.0688158203125002
11 0.0547997070312502
12 0.0433041015625001
13 0.0339080078125
14 0.0281607421874999
15 0.0210115234374999
16 0.0159416015625
17 0.013575390625
18 0.0101943359375
};
\addlegendentry{Random}
\addplot [semithick, red, dashed]
table {%
0 0.5
1 0.3201904296875
2 0.20880224609375
3 0.155491015625001
4 0.118972656250001
5 0.0846564453125002
6 0.0683878906250001
7 0.0581800781250001
8 0.04395986328125
9 0.0338941406249999
10 0.0278213867187499
11 0.0230004882812499
12 0.01899560546875
13 0.0150919921875
14 0.0135923828125
15 0.0116212890625
16 0.01030810546875
17 0.0101529296875
18 0.0101943359375
};
\addlegendentry{Bayesian Entropy}
\addplot [semithick, blue, dashed]
table {%
0 0.5
1 0.317187499999997
2 0.218678808593748
3 0.172235253906248
4 0.123334277343749
5 0.0972216796875
6 0.0832712890624999
7 0.0641038085937499
8 0.0548044921874999
9 0.04663076171875
10 0.03955947265625
11 0.0320949218749999
12 0.0270456054687499
13 0.0228884765625
14 0.0172658203125
15 0.01367705078125
16 0.0123193359375
17 0.01206787109375
18 0.0101943359375
};
\addlegendentry{Bayesian Mode}
\end{axis}

\end{tikzpicture}
\caption{Accuracy (left) and Brier distance (right) of TAs for a single-skill BN/CN}
\label{fig:mono}
\end{figure}

For a very first validation of our approach, we consider a simple setup made of a single Boolean skill $S$ and a repository with 18 Boolean questions based on nine different parametrizations (two questions for parametrization). In such BN, the CPT of a question can be parametrized by two numbers. E.g., in the example in Figure \ref{fig:minicat}, we used the probabilities of correctly answering the question given that the skill is present or not, i.e., $P(Q=1|S=1)$ and $P(Q=1|S=0)$. A more interpretable parametrization can be obtained as follows:
\begin{eqnarray}
\delta &:=& 1-\frac{1}{2}[P(Q=1|S=1)+P(Q=1|S=0)]\,,\\
\kappa &:=& P(Q=1|S=1)-P(Q=1|S=0)\,.
\end{eqnarray}
Note that $P(Q=1|S=1)>P(Q=1|S=0)$ is an obvious rationality constraint for questions, otherwise having the skill would make less likely to answer properly to a question. Both parameters are therefore non-negative. Parameter $\delta$, corresponding to the (arithmetic) average of the probability for a wrong answer over the different skill values, can be regarded as a normalized index of the question \emph{difficulty}. E.g., in Figure \ref{fig:minicat}, $Q_1$ ($\delta=0.4$) is less difficult than $Q_2$ ($\delta=0.5$). Parameter $\kappa$ can be instead regarded as a descriptor of the difference of the conditional PMFs associated with the different skill values. In the most extreme case $\kappa=1$, the CPT $P(Q|S)$ is diagonal implementing an identity mapping between the skill and the question. We therefore regard $\kappa$ as a indicator of the \emph{discriminative} power of the question. In our tests, for the BN quantification, we consider the nine possible parametrizations corresponding to $(\delta,\gamma) \in [0.4,0.5,0.6]^2$. For $P(S)$ we use instead a uniform quantification. For the CN approach we perturb all the BN parameters with $\epsilon=\pm 0.05$, thus obtaining a CN quantification. A group of 1024 simulated students, half of them having $S=0$ and half with $S=1$ is used for simulations. The student answers are sampled from the CPT of the asked question on the basis of the student profile. Figure \ref{fig:mono} (left) depicts the accuracy of the BN and CN approaches based on both the entropy and the DM scores. For credal models, decisions are based on the mid-point between the lower and the upper probability, while lower entropy and conditional entropies are used. We notably see all the adaptive approaches outperforming a non-adaptive, random, choice of the questions. To better investigate the strong overlap between these trajectories, in Figure \ref{fig:mono} (right) we compute the Brier score and we might observe the strong similarity between DM and entropy approaches in both the Bayesian and the credal case, with the credal approaches slightly outperforming the Bayesian ones.

\subsection{Multi-Skill Experiments on Real Data}
For a validation on real data, we consider an online German language placement test (see also \cite{mangili2017b}). Four different Boolean skills associated with different abilities (vocabulary, communication, listening and reading) are considered and modeled by a chain-shaped graph, for which BN and CN quantification are already available. A repository of 64 Boolean questions, 16 for each skill, with four different levels of difficulty and discriminative power, have been used.

Experiments have been achieved by means of the CREMA library for credal networks \cite{huber2020a}.\footnote{\url{github.com/IDSIA/crema}} The Java code used for the simulations is available together with the Python scripts used to analyze the results and the model specifications.\footnote{\url{github.com/IDSIA/adaptive-tests}}

\begin{figure}[htp!]
\centering
\begin{tikzpicture}[scale=0.6]
\begin{axis}[
legend cell align={left},
legend style={
  fill opacity=0.8,
  draw opacity=1,
  text opacity=1,
  at={(0.97,0.03)},
  anchor=south east,
  draw=white!80!black
},
tick align=outside,
tick pos=both,
x grid style={white!69.0196078431373!black},
xlabel={Number of questions},
xmin=-3.2, xmax=67.2,
xtick style={color=black},
y grid style={white!69.0196078431373!black},
ylabel={Aggregated Accuracy},
ymin=-0.05, ymax=1.05,
ytick style={color=black}
]
\addplot [semithick, red]
table {%
0 0
1 0.80078125
2 0.8671875
3 0.89453125
4 0.921875
5 0.953125
6 0.9609375
7 0.97265625
8 0.9765625
9 0.984375
10 0.99609375
11 0.9921875
12 0.9921875
13 0.9921875
14 1
15 0.99609375
16 0.9921875
17 0.99609375
18 0.99609375
19 0.99609375
20 0.99609375
21 1
22 0.99609375
23 0.99609375
24 0.99609375
25 0.99609375
26 0.99609375
27 0.99609375
28 0.99609375
29 0.99609375
30 0.99609375
31 0.99609375
32 0.99609375
33 0.99609375
34 0.99609375
35 0.99609375
36 0.99609375
37 0.99609375
38 0.99609375
39 0.99609375
40 0.99609375
41 0.99609375
42 0.99609375
43 0.99609375
44 0.99609375
45 0.99609375
46 0.99609375
47 0.99609375
48 0.99609375
49 0.99609375
50 0.99609375
51 0.99609375
52 0.99609375
53 0.99609375
54 0.99609375
55 0.99609375
56 0.99609375
57 0.99609375
58 0.99609375
59 0.99609375
60 0.99609375
61 0.99609375
62 0.99609375
63 0.99609375
64 0.99609375
};
\addlegendentry{Credal Entropy}
\addplot [semithick, blue]
table {%
0 0
1 0.89453125
2 0.859375
3 0.86328125
4 0.88671875
5 0.89453125
6 0.93359375
7 0.95703125
8 0.9609375
9 0.97265625
10 0.9765625
11 0.9765625
12 0.97265625
13 0.97265625
14 0.97265625
15 0.98046875
16 0.984375
17 0.984375
18 0.98828125
19 0.98828125
20 0.98828125
21 0.98828125
22 0.98828125
23 0.98828125
24 0.98828125
25 0.98828125
26 0.98828125
27 0.98828125
28 0.98828125
29 0.98828125
30 0.98828125
31 0.98828125
32 0.98828125
33 0.98828125
34 0.98828125
35 0.98828125
36 0.98828125
37 0.98828125
38 0.98828125
39 0.98828125
40 0.98828125
41 0.98828125
42 0.98828125
43 0.98828125
44 0.98828125
45 0.98828125
46 0.98828125
47 0.98828125
48 0.98828125
49 0.98828125
50 0.98828125
51 0.98828125
52 0.98828125
53 0.98828125
54 0.98828125
55 0.98828125
56 0.98828125
57 0.98828125
58 0.98828125
59 0.98828125
60 0.98828125
61 0.98828125
62 0.98828125
63 0.98828125
64 0.98828125
};
\addlegendentry{Credal Mode}
\addplot [semithick, black, dashed]
table {%
0 0
1 0.640625
2 0.640625
3 0.6484375
4 0.8203125
5 0.7421875
6 0.78125
7 0.82421875
8 0.83203125
9 0.83984375
10 0.8671875
11 0.859375
12 0.8671875
13 0.87890625
14 0.875
15 0.875
16 0.875
17 0.91015625
18 0.90234375
19 0.8984375
20 0.90625
21 0.921875
22 0.94140625
23 0.9296875
24 0.92578125
25 0.9375
26 0.94140625
27 0.94921875
28 0.9453125
29 0.94140625
30 0.94140625
31 0.94921875
32 0.94921875
33 0.94140625
34 0.9453125
35 0.95703125
36 0.9609375
37 0.96484375
38 0.95703125
39 0.953125
40 0.94921875
41 0.953125
42 0.953125
43 0.95703125
44 0.95703125
45 0.96875
46 0.96875
47 0.96875
48 0.97265625
49 0.97265625
50 0.97265625
51 0.97265625
52 0.9765625
53 0.9765625
54 0.9765625
55 0.9765625
56 0.9765625
57 0.98046875
58 0.98828125
59 0.98828125
60 0.98828125
61 0.98828125
62 0.9921875
63 0.98828125
64 0.98828125
};
\addlegendentry{Random}
\addplot [semithick, red, dashed]
table {%
0 0
1 0.6171875
2 0.71875
3 0.78515625
4 0.8046875
5 0.8671875
6 0.87890625
7 0.89453125
8 0.91796875
9 0.921875
10 0.92578125
11 0.93359375
12 0.9609375
13 0.9609375
14 0.95703125
15 0.96484375
16 0.96484375
17 0.96875
18 0.97265625
19 0.97265625
20 0.97265625
21 0.97265625
22 0.9765625
23 0.97265625
24 0.984375
25 0.984375
26 0.98046875
27 0.984375
28 0.984375
29 0.984375
30 0.98828125
31 0.98828125
32 0.98828125
33 0.98828125
34 0.98828125
35 0.984375
36 0.984375
37 0.98828125
38 0.98828125
39 0.98828125
40 0.98828125
41 0.98828125
42 0.98828125
43 0.98828125
44 0.98828125
45 0.98828125
46 0.98828125
47 0.98828125
48 0.98828125
49 0.98828125
50 0.98828125
51 0.98828125
52 0.98828125
53 0.98828125
54 0.98828125
55 0.98828125
56 0.98828125
57 0.98828125
58 0.98828125
59 0.98828125
60 0.98828125
61 0.98828125
62 0.98828125
63 0.98828125
64 0.98828125
};
\addlegendentry{Bayesian Entropy}
\addplot [semithick, blue, dashed]
table {%
0 0
1 0.6171875
2 0.71875
3 0.78515625
4 0.796875
5 0.84375
6 0.875
7 0.890625
8 0.90234375
9 0.91015625
10 0.9140625
11 0.93359375
12 0.94921875
13 0.9453125
14 0.96484375
15 0.95703125
16 0.9609375
17 0.96875
18 0.97265625
19 0.96484375
20 0.96484375
21 0.96875
22 0.96484375
23 0.96484375
24 0.96484375
25 0.9609375
26 0.953125
27 0.95703125
28 0.95703125
29 0.96484375
30 0.96484375
31 0.96484375
32 0.96875
33 0.96875
34 0.96875
35 0.96875
36 0.96875
37 0.96875
38 0.97265625
39 0.9765625
40 0.97265625
41 0.97265625
42 0.97265625
43 0.97265625
44 0.97265625
45 0.97265625
46 0.97265625
47 0.96875
48 0.97265625
49 0.9765625
50 0.9765625
51 0.9765625
52 0.98046875
53 0.984375
54 0.98046875
55 0.984375
56 0.984375
57 0.984375
58 0.98828125
59 0.98828125
60 0.98828125
61 0.98828125
62 0.98828125
63 0.98828125
64 0.98828125
};
\addlegendentry{Bayesian Mode}
\end{axis}

\end{tikzpicture}
\caption{Aggregated Accuracy for a multi-skill TA}
\label{fig:multi}
\end{figure}
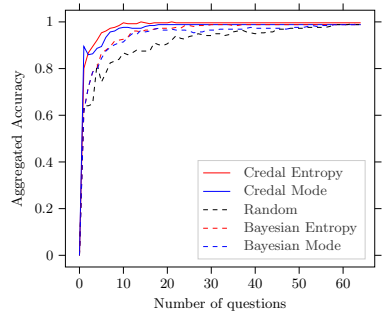

Performances are evaluated as for the previous model, the only difference being that here the accuracy is aggregated by average over the separate accuracies for the four skills. The observed behaviour, depicted in Figure \ref{fig:multi}, is analogous to that of the single skill case: entropy-based and mode-based scores are providing similar results, with the credal approach typically leading to more accurate evaluations (or evaluations of the same quality with fewer questions).

\section{Outlooks and Conclusions}\label{sec:conc}
A new score for adaptive testing in Bayesian and credal networks has been proposed. Our proposal is based on indexes of qualitative variation, being in particular focused on the modal probability for their explainability features. An algorithm to evaluate this quantity in the credal case is derived. Our experiments show that moving to these scores does not really affect the quality of the selection process. Besides a deeper experimental validation, a necessary future work consists in the derivation of simpler elicitation strategies for these model in order to promote their application to real-world testing environments.  
\bibliographystyle{splncs04}
\bibliography{paper}
\end{document}